\documentclass[letterpaper]{article} 
\usepackage{aaai25}  
\usepackage{times}  
\usepackage{helvet}  
\usepackage{courier}  
\usepackage[hyphens]{url}  
\usepackage{graphicx} 
\def\UrlFont{\rm}  
\usepackage{natbib}  
\usepackage{caption} 
\usepackage{algorithm}
\usepackage{algorithmic}

\usepackage{amsmath}
\usepackage{amsfonts}
\usepackage{amssymb}

\usepackage{booktabs}

\DeclareMathOperator*{\argmax}{arg\,max}

\newtheorem{proposition}{Proposition}
\newtheorem{assumption}{Assumption}

\usepackage{newfloat}
\usepackage{listings}
\DeclareCaptionStyle{ruled}{labelfont=normalfont,labelsep=colon,strut=off} 
\floatstyle{ruled}
\newfloat{listing}{tb}{lst}{}
\floatname{listing}{Listing}
\title{Behavior Preference Regression for Offline Reinforcement Learning}
\author{
    Padmanaba Srinivasan,
    William Knottenbelt
}
\affiliations{
    Department of Computing, Imperial College London\\
    \{ps3416, wjk\}@imperial.ac.uk
}

\begin{document}

\maketitle

\begin{abstract}
    Offline reinforcement learning (RL) methods aim to learn optimal policies with access only to trajectories in a fixed dataset. Policy constraint methods formulate policy learning as an optimization problem that balances maximizing reward with minimizing deviation from the behavior policy. Closed form solutions to this problem can be derived as weighted behavioral cloning objectives that, in theory, must compute an intractable partition function. Reinforcement learning has gained popularity in language modeling to align models with human preferences; some recent works consider \textit{paired} completions that are ranked by a preference model following which the likelihood of the preferred completion is directly increased. We adapt this approach of paired comparison. By reformulating the paired-sample optimization problem, we \textit{fit} the maximum-mode of the Q function while maximizing \textit{behavioral consistency} of policy actions. This yields our algorithm, \underline{B}ehavior \underline{P}reference \underline{R}egression for offline RL (BPR). We empirically evaluate BPR on the widely used D4RL Locomotion and Antmaze datasets, as well as the more challenging V-D4RL suite, which operates in image-based state spaces. BPR demonstrates state-of-the-art performance over all domains. Our on-policy experiments suggest that BPR takes advantage of the stability of on-policy value functions with minimal perceptible performance degradation on Locomotion datasets.
\end{abstract}

%

\section{Introduction}

As reinforcement learning (RL) sees increasing application in a variety of fields, from control \citep{RN964} to language modeling \citep{RN965}, it has also become increasingly \textit{data-hungry} \citep{RN967}. The need to acquire data through \textit{online} interaction can make deep reinforcement learning infeasible in many domains. In response, one direction of research develops \textit{offline} RL algorithms that aim to learn from a static dataset of pre-collected interactions \citep{RN695}. 

Standard off-policy algorithms can be directly applied on offline datasets \citep{RN802,RN841}, though in practice the combined effect of off-policy learning, bootstrapping, and function approximation \citep{RN679} introduces extrapolation error. The resulting distribution shift between the learned policy and behavior policy can cause training instability and subsequent failure when deployed in the real environment \citep{RN684}.

Offline RL algorithms address the challenges of offline off-policy evaluation in one of three ways: 1) incorporating pessimism into value estimation, 2) imposing policy constraints or 3) avoiding off-policy evaluation altogether by learning an on-policy value function. Pessimism offers performance guarantees \citep{RN782}, policy constraints may make better use of the representational power of neural networks \citep{RN979} and learning on-policy values is more stable and avoids the overestimation and iterative exploitation associated with off-policy evaluation \citep{RN882}.

Another approach to learning aims to align policy rollouts with human preferences \citep{RN983,RN984,RN965}. Preference-based RL is popular in language modeling under the banner of RL from human feedback. Recent methods take models trained using supervised learning and finetune them using an offline dataset by directly increasing the likelihood of generating preferred sequences \citep{RN936}. 

The principle of aligning policies with human preferences has been explored in offline RL \citep{RN939,RN936,RN938}. While they aim to solve the same tasks, preference-based methods must either directly learn to generate aligned sequences \citep{RN939} or must train a preference model \citep{RN936,RN938} on specially crafted datasets of human preferences. These methods typically eschew more traditional reward modeling (RM) and perform in-sample learning using pairs of trajectories. 

\paragraph{Contributions} Motivated by finetuning approaches to align language models \citep{RN936,RN942}, in this work we develop a policy objective for offline RL that directly learns the policy density: our algorithm performs \underline{B}ehavior \underline{P}reference \underline{R}egression for offline RL (BPR). We analyze BPR with respect to regularized value functions in the context of preference models to demonstrate theoretical performance improvement. Evaluation on D4RL \citep{RN731} demonstrates that BPR achieves SOTA performance on Locomotion and Antmaze datasets. Additional tests on the image-based V-D4RL \citep{RN1000} tasks reveal that BPR is able to transition across modalities to achieve high performance in non-proprioceptive domains. In experiments with on-policy value functions, BPR outperforms competing methods by a substantial margin on four of six datasets. By incorporating more expressive ensembles of value functions, BPR improves performance substantially on tasks that typically require trajectory stitching.

\section{Related Work}

Reinforcement learning aims to solve sequential decision-making tasks formulated as a Markov Decision Process (MDP), $\mathcal{M} = \{\mathcal{S}, \mathcal{A}, \mathcal{R}, P, p_0, \gamma\}$, where $\mathcal{S}$ denotes the state space, $\mathcal{A}$ the action space, $\mathcal{R}$ a scalar reward function, $P$ the transition dynamics, $p_0$ the initial state distribution, and $\gamma \in [0, 1)$ the discount factor. The goal of RL is to learn an optimal policy that executes actions such that it maximizes the expected discounted reward; for any policy $\pi$ we denote its return as $\eta(\pi) = \mathbb{E}_{\tau \sim \rho_{\pi}(\tau)} \left[ \sum_{t=0}^{T} \gamma^t \mathcal{R}(s_t, a_t) \right]$ where $\rho_{\pi} (\tau) = p_0 (s_0) \prod_{t=1}^{T} \pi(a_t | s_t) P (s_{t+1} | s_t, a_t)$ is a trajectory sampled under policy $\pi$ \citep{RN679}.

\subsection{Offline Reinforcement Learning}

Offline RL methods aim to maximize sample efficiency and learn optimal policies given only a static dataset of interactions $\mathcal{D} = \{s, a, r, s\}_{n=1}^{N}$, which was produced by one or more unknown behavior policies of uncertain quality. 

The tuples that form the dataset contain information that we are certain about. Actions beyond the support of the dataset are of unknown quality and lead to unknown trajectories. Generally, offline RL methods aim to train policies that maximize expected reward while remaining within the dataset support. 

\paragraph{Off-Policy Methods} A large body of offline RL methods adapt existing off-policy algorithms for the offline domain. Approaches can be classified into those that apply critic regularization to address overestimation and those that impose policy constraints to draw the current policy towards the dataset support.

Critic regularizers can explicitly reduce the values of OOD actions \citep{RN698,RN918}, thus shaping the Q function. This forces the policy to maximize Q values that are in-support. Regularizers can function implicitly by making use of the diversity-based-pessimism of large ensembles of value functions \citep{RN770,RN771}. Ensembles condone some degree of OOD action selection which \citet{RN770} attribute to improving performance. \citet{RN813} explore this further and find that relaxing constraints can improve performance in algorithms without large ensembles. Work by \citet{RN771} suggests that large min-clipped ensembles may be redundant due to the collapse in independence of ensemble members.

Policy constraints aim to directly confine the actor to select in-support actions. These are typically formulated explicitly as divergence penalties \citep{RN699,RN719}, implicitly through weighted behavioral cloning (BC) \citep{RN708,RN705,RN706} or by architecturally limiting the exploration afforded to the policy \citep{RN684}. 

\paragraph{On-Policy Methods} \citet{RN882} recognize off-policy evaluation as a source of instability in offline RL and instead learn an on-policy (Onestep) value function. The policy learned using this value function outperforms those learned via behavioral cloning and some offline off-policy methods. On-policy learning is extended by \citet{RN711} and \citet{RN954} who attempt to approximate the in-sample maximum return by dataset trajectories which they use to train a weighted BC policy. \citet{RN943} adapt online, on-policy PPO \citep{RN982} for the offline setting and develop an algorithm that uses offline datasets with periodic online evaluation. This is not a fully offline RL algorithm and their own experiments show that without online evaluation to enable policy replacement, performance will degrade.

\subsection{Preference-Based Reinforcement Learning}

Building on the ideas of \citet{RN983} and \citet{RN984}, \citet{RN965} suggest using preference-annotated data as reward signals to train language models that are better aligned with human values. Subsequent work has developed learning from preferences further \citep{RN985} with the notable \textit{Direct Preference Optimization} (DPO) \citep{RN936} which finetunes a maximum likelihood trained policy on an offline dataset of paired preference annotated data by directly optimizing policy density as a proxy for the reward function. 

In continuous-control offline RL, \citet{RN939} train a trajectory-producing policy on non-Markovian, preference-based rewards. \citet{RN1115} use a preference labeled dataset to train a preference model that is subsequently used to label preferred trajectories in an unlabeled dataset used for policy training. Using preference datasets, \citet{RN938} directly train a policy (similar to DPO) as an optimal advantage function using preference data. Common themes of preference-based offline RL methods are the eschewing of traditional rewards for human-annotated data, and the requirement trajectories to be \textit{paired} for preference learning which does not allow evaluation of OOD actions.

\section{Behavior Preference Regression}

We consider the general, reverse KL-constrained problem:
\begin{align}
    \label{eq: KL constrained problem}
    \nonumber \pi_{t+1} = \argmax_{\pi \in \Pi} &\mathbb{E}_{s \sim \mathcal{D}, a \sim \pi} [ f(s, a) 
    \\
    &- \lambda D_{\text{KL}} (\pi (\cdot | s) || \pi_{\text{ref}} (\cdot | s)) ],
\end{align}
\noindent where $\lambda \geq 0$ controls the tradeoff between remaining close to a distribution $\pi_{\text{ref}}$ and maximizing some function $f(\cdot, \cdot)$.

The closed form solution to the optimization problem has been previously derived \citep{RN935,RN987}:
\begin{align}
    \label{eq: KL constrained problem solution}
    \pi_{t+1} = \pi_{\text{ref}}(a | s) \exp (\frac{1}{\lambda} f(s, a)) \frac{1}{Z(s)}
    \\
    Z(s) = \int_{a \in \mathcal{A}} \pi_{\text{ref}} (a | s) \exp(\frac{1}{\lambda} f(s, a))\, da,
\end{align}
\noindent where $Z(s)$ is the partition function. 

Using the \textit{DPO trick} \citep{RN936}, we can rearrange Equation~\ref{eq: KL constrained problem solution} as:
\begin{align}
    f(s, a) = \lambda \left( \log Z(s) + \log \frac{\pi_{t+1} (a | s)}{\pi_{\text{ref}} (a | s)} \right),
\end{align}

\noindent following which using ranked, paired samples where $a_1 \succ a_2$ we can write:
\begin{align}
    \label{eq: paired equality}
    \nonumber &f(s, a_1) - f(s, a_2) = 
    \\ 
    &\quad\lambda \left( \log \frac{\pi_{t+1} (a_1 | s)}{\pi_{\text{ref}} (a_1 | s)} - \log \frac{\pi_{t+1} (a_2 | s)}{\pi_{\text{ref}} (a_2 | s)} \right),
\end{align}
\noindent which conveniently cancels out the partition function. 

DPO takes the binary preference $a_1 \succ a_2$ and passes the RHS through a Bradley-Terry preference model \citep{RN937} to optimize for $a_1$. Consequently, DPO fails to capture \textit{how much more} $a_1$ is preferred to $a_2$. \citet{RN942} aim to directly learn the relative difference by solving the regression problem:
\begin{align}
    \label{eq: policy regression objective}
    \nonumber &\bigl[ \left( f(s, a_1) - f(s, a_2) \right) -  
    \\
    &\quad\lambda \left( \log \frac{\pi_{t+1} (a_1 | s)}{\pi_{\text{ref}} (a_1 | s)} - \log \frac{\pi_{t+1} (a_2 | s)}{\pi_{\text{ref}} (a_2 | s)} \right) \bigr]^2.
\end{align}

In this work, we focus on learning a policy by solving this relative regression problem. 

\subsection{What do we \textit{Prefer} in Offline RL?} 

Most policy constraint formulations typically choose ${f(s, \cdot) = Q(s, \cdot)}$ and ${\pi_{\text{ref}} (\cdot | s) = \hat{\pi}_{\beta} (\cdot | s)}$ where $\hat{\pi}_{\beta}$ is an empirical behavior policy. This follows the principle of maximizing reward while satisfying some constraint that must be carefully balanced by tuning $\lambda$ to curb the distribution shift \citep{RN882}. 

We propose an alternative optimization: we \textbf{maximize behavioral consistency} and \textbf{reverse KL fit the (maximum) mode of the Q function}~-- in preference terms, we fit a distribution of high-reward actions and regress toward actions with high likelihood under the behavior policy.

\paragraph{Selecting $\mathbf{\pi_{\text{ref}}}$} Soft Q-learning \citep{RN802} trains a maximum entropy Q function that can be written as an energy-based model (EBM) \citep{RN884}. We formulate ${\pi_{\text{ref}} (a | s) = \frac{\exp (Q(s, a))}{Z_{Q}(s)}}$ where ${{Z_{Q}(s) = \int_{\mathcal{A}} Q(s, a)\, da}}$ is the partition function, which subsequently cancels out in the RHS of Equation~\ref{eq: policy regression objective}. This allows us to directly optimize the soft actor--critic (SAC) policy objective \citep{RN802} without resorting to approximations of the entropy through a \texttt{tanh}-transformed Gaussian. 

\paragraph{Selecting $f(\cdot, \cdot)$} The true behavior policy is unknown and so we must make an empirical approximation. Prior methods typically learn explicit policies using behavioral cloning \citep{RN918,RN699,RN943}. This can be limiting, as the number of behavior policy modes must be known beforehand. Implicit policies offer more flexible behavior models \citep{RN885}. We train an implicit behavior policy $\hat{\pi}_{\beta}$ as an EBM that learns an energy function ${E(s, a) \in \mathbb{R}}$. We recover an estimate of the explicit behavior policy using the Boltzmann distribution:
$ \hat{\pi}_{\beta} (a | s) = \frac{\exp (-E(s, a))}{Z_{E} (s)}$ where $Z_{E} (s) = \int_{\mathcal{A}} \exp (-E(s, a))\, da$ is the EBM partition function. 

Fortunately, $Z_{E} (s)$ also cancels out when using $f(s, \cdot) = \log \hat{\pi}_{\beta}(\cdot | s)$ in the LHS of Equation~\ref{eq: policy regression objective} and we only need to compute $E(s, a_1)$ and $E(s, a_2)$. Using an EBM behavior policy, we make no \textit{inductive bias} with respect to the (multi)modality of the true behavior policy.

Combining everything, our policy optimization objective is:
\begin{align}
    \label{eq: our policy regression objective}
    \nonumber &[ \left( E(s, a_2)) - E(s, a_1) \right) -  
    \\
    &\quad\lambda \left( \log \frac{\pi_{t+1} (a_1 | s)}{\exp(Q(s, a_1))} - \log \frac{\pi_{t+1} (a_2 | s)}{\exp(Q(s, a_2))} \right) ]^2.
\end{align}

\paragraph{Interpretation} Learning $\pi^*$ requires a policy to select in-sample actions that also maximize expected reward. By selecting the regression target to be the difference ${\log \hat{\pi}_{\beta} (a_1 | s) - \log \hat{\pi}_{\beta} (a_2 | s)}$, we treat the behavior EBM as an expert preference model that communicates by how much $a_1 \succ a_2$. This differs from previous preference-based offline RL formulations that evaluate the preference by comparing discounted rewards over entire trajectories (produced by the behavior policy) for a pair of actions. Such reward-based preference learning has been shown to be inconsistent with human-preference labels \citep{RN992}. Placing a support constraint on the policy towards high-reward modes in the soft Q function and combining this with off-policy evaluation offers a far more flexible approach without the need for human-labeled preference datasets. Most importantly, we \textbf{never} need to compute any partition function $Z(s), Z_{Q} (s)$ or $Z_{E} (s)$~-- past work has found that approximating partition functions, though technically correct, is deleterious to performance \citep{RN706}.

\subsection{Self-Play} 

Let $\mu_1, \mu_2$ be the sampling distributions for $a_1$ and $a_2$, respectively. Offline preference-based methods use datasets that contain previously evaluated pairs of completions sampled from $\pi_{\beta}$ \citep{RN939,RN936,RN938}. In standard offline settings, samples are drawn from $\mathcal{D}$ or $\pi$ and in the paired setting this equates to using $\mu_1 = \pi_{\beta} = \mathcal{D}$ and $\mu_2 = \pi$ (reference sampling). Recently, \citet{RN989} prove that performing self-play with multiple samples drawn from $\pi$ itself results in stable learning with strong theoretical guarantees~-- this involves sampling a pair of actions from the current policy and querying a learned preference/reward model to optimize Equation~\ref{eq: KL constrained problem solution}. We use self-play to sample actions for policy optimization, hence $\mu_1 = \mu_2 = \pi$. We compare reference sampling and self-play schemes in a toy bandit example in the Appendix.

\subsection{Analysis} 

Rearranging Equation~\ref{eq: paired equality} and inserting $\pi_{\text{ref}}(\cdot | s) = \frac{\exp (Q(s, \cdot))}{Z_Q(s)} $ and $f(s, \cdot) = \log {\pi}_{\beta}(\cdot | s)$, we obtain:
\begin{align}
    \label{eq: implicit Q value formulation}
    \nonumber(Q(s, a_1) + \frac{1}{\lambda} &\log {\pi}_{\beta}(a_1 | s)) - 
    \\
    \nonumber(Q(s, a_2) &+ \frac{1}{\lambda} \log {\pi}_{\beta} (a_2 | s))
    \\ 
    &= \log \pi_{t+1} (a_1 | s) - \log \pi_{t+1} (a_2 | s).
\end{align}

We explicitly cancel $Z_Q (s)$ but leave $Z_E (s)$ unfactorized for clarity. 

We define $\tilde{Q}(s, a) \triangleq Q(s, a) + \frac{1}{\lambda} \log \hat{\pi}_{\beta} (a | s)$ and notice that this is a variation of an implicit Q function popular in \textit{online} RL \citep{RN956,RN993} and is exactly the Q function formulation used by Fisher-BRC when using $\lambda = 1.0$ \citep{RN918}. We subsequently interpret that our policy regression objective is equivalent to fitting the policy to the implicit Q function. 

We rewrite the LHS of Equation~\ref{eq: implicit Q value formulation} as a \textit{soft} preference function:
\begin{align}
    \texttt{P} (s, a_1, a_2) \triangleq \tilde{Q} (s, a_1) - \tilde{Q} (s, a_2).
\end{align}

\begin{assumption}{(Tuned Preference Function)}
    \label{assumption: tuned preferences}
    \begin{align}
        \nonumber \texttt{P} (s, a_1, a_2) \geq 0\quad \forall a_1, a_2 \in \mathcal{A} 
        \\
        \nonumber \text{when}\quad \pi_{\beta} (a_1 | s) \geq \pi_{\beta} (a_2 | s).
    \end{align}
\end{assumption}

We assume that any action $a_1$ with a higher likelihood under the behavior policy than $a_2$ is preferred. In practice, this can be satisfied by tuning $\lambda$. 

For any policy $\pi$, recall its return is given by $\eta(\pi) = \mathbb{E}_{\tau \sim \rho_{\pi}(\tau)} \left[ \sum_{t=0}^{T} \gamma^t \mathcal{R}(s_t, a_t) \right]$. The behavior policy used to produce the dataset is $\pi_{\beta}$ and let the policy learned by optimizing using $\tilde{Q}(s, a)$ be $\tilde{\pi}$ (i.e.\ the policy that maximizes soft preferences). 

\begin{proposition}{(Perfect Preference Model)}
    If the preference function $\texttt{P}(s, a_1, a_2)$ is perfect i.e.\ $\tilde{Q}^* = Q^* + \pi_{\beta}$ is accurate, then the deterministic policies $\pi_{\beta}$ and $\tilde{\pi}$ satisfy:
    \begin{align}
        \nonumber &\eta (\tilde{\pi}) - \eta (\pi_{\beta}) 
        \\
        &\quad\approx 
        \mathbb{E}_{s \sim \mathcal{D}} \left[ \tilde{Q}^* (s, \tilde{\pi} (s)) - \tilde{Q}^* (s, \pi_{\beta} (s)) \right] \geq 0
    \end{align}
\end{proposition}

In practice, estimation is noisy. For $\tilde{Q}$, this comes from two sources: errors are present in both Q function and behavior policy estimates. EBM approximation error has been studied by \citet{RN885} (Theorem~2) who prove that a Lipschitz-continuous EBM policy can exhibit arbitrarily small error. 

The total variational distance between two value functions $Q_1, Q_2$ is: $D_{\text{TV}}(Q_1, Q_2) = \max_{s \in \mathcal{S}} \lvert Q_1(s, \pi (s)) - Q_2 (s, \pi(s)) \rvert$.

\begin{proposition}{(Noisy Preference Model)}
    Consider the case where $\hat{\pi}_{\beta}$ and $Q^*$ contain errors and produce the noisy $\tilde{Q}^{-}$. Then $\forall \tilde{Q}^{-}$ where $D_{\text{TV}} (\tilde{Q}^{-} (s, \tilde{\pi}(s)), Q^* (s, \tilde{\pi}(s))) \leq \tilde{\epsilon}$ and $D_{\text{TV}} (\tilde{Q}^{-} (s, \pi_{\beta}(s)), Q^{*} (s, \pi_{\beta}(s))) \leq \epsilon$ the following holds:
    \begin{align}
        \nonumber \eta (\tilde{\pi}) &- \eta (\pi_{\beta})
        \\
        \nonumber &\leq \mathbb{E}_{s \sim \mathcal{D}} \left[ \tilde{Q}^{-}(s, \tilde{\pi} (s)) - \tilde{Q}^{-} (s, \pi_{\beta} (s)) \right] 
        \\
        &\quad+ 2 \rho_{\text{max}} (\tilde{\epsilon} + \epsilon)
        \\
        \nonumber &\text{where}\quad \rho_{\text{max}} = \sup \{ \rho_{\pi_{\beta}}(s), s \in \mathcal{S}\}.
    \end{align}
\end{proposition}

We defer proofs to the Appendix.

The first term after the inequality is non-negative under Assumption~\ref{assumption: tuned preferences} and the second term is present due to the modeling error of the estimated Q function and behavior policy. This can be reduced by using a more accurate function approximator.

\subsection{Implementation}

Our actor--critic implementation follows a standard implementation of SAC \citep{RN802} with modifications to the policy improvement step. We illustrate our policy improvement in Algorithm~\ref{algo: policy improvement} and provide additional implementation details in the Appendix. 

\begin{algorithm}[tb]
    \caption{Policy improvement step. Comment NG denotes steps where gradients do not have to be computed.}
    \label{algo: policy improvement}
    \textbf{Require}: Offline dataset $\mathcal{D}$, pretrained EBM $E(\cdot, \cdot)$, training steps $N$\\
    \textbf{Output}: Trained policy $\pi$ 
    \begin{algorithmic}[0] 
        \STATE Let $t=0$.
        \FOR{{t = 1} \TO {$N$}}
        \STATE Sample $(s, a, r, s') \sim \mathcal{D}$
        \STATE Sample $a_1, a_2 \sim \pi$ $\quad \#$ NG
        \STATE Compute $\log \pi(a_1 | s), \log \pi(a_2 | s)$ 
        \STATE Compute $E(s, a_1) \text{ and } E(s, a_2)$ $\quad \#$ NG
        \STATE Compute $Q(s, a_1) \text{ and } Q(s, a_2)$ $\quad \#$ NG
        \STATE Update $\pi$ using Equation~\ref{eq: our policy regression objective}. 
        \STATE $\#$ Update critics
        \ENDFOR
    \STATE \textbf{return} $\pi$
    \end{algorithmic}
\end{algorithm}

The EBM approximation of $\pi_{\beta}$ is trained prior to the main actor--critic training phase. We follow design decisions detailed in \citet{RN885}, using spectral normalization \citep{RN888} and deep networks. 

\paragraph{Summary of Hyperparameters} In addition to the standard hyperparameters of SAC (clipped double-Q learning \citep{RN714}, entropy regularized off-policy Q functions), our algorithm introduces the hyperparameter $\lambda$, which controls the tradeoff between the KL constraint and maximizing behavioral consistency. 

In general, we find that simply \textbf{using $\lambda = 1.0$ works well across all tasks}; our primary results use this hyperparameter value and we perform ablations to evaluate sensitivity in our experiments.

\section{Experiments}

In this section, we evaluate empirically BPR and aim to answer the following questions:
\begin{itemize}
    \item How well does BPR perform compared to state-of-the-art offline RL methods?
    \item Does BPR perform well in tasks with visual state spaces?
    \item Can Onestep-trained policies compete with off-policy off\-line RL?
    \item How sensitive is BPR to values of $\lambda$?
\end{itemize}

\paragraph{Experimental Setup} In all BPR experiments, we report the normalized mean score with standard deviation on five seeds over 100 evaluations in Antmaze tasks and 10 in others. All scores are reported using the policy from the final checkpoint.

\paragraph{Baselines} We compare results against the following, well-known baselines: CQL \citep{RN698}, IQL \citep{RN711} and TD3+BC \citep{RN719}. We also include the recent offline RL algorithms: ReBRAC \citep{RN842}, XQL \citep{RN954} and Diff-QL \citep{RN999} (which replaces a Gaussian/deterministic policy with a Diffusion policy \citep{RN623}). Of the latter three methods, both ReBRAC and XQL tune hyperparameters extensively for each dataset. In contrast, the older baselines, Diff-QL and our BPR find hyperparameters that generalize well across like-tasks (i.e.\ the same hyperparameters for all Locomotion tasks etc.).

For a more comprehensive comparison, we also include the preference-based offline RL methods: PT \citep{RN939}, OPPO \citep{RN997} and DPPO \citep{RN1115}. 

\subsection{D4RL}

We evaluate BPR on D4RL Locomotion and Antmaze datasets \citep{RN731}. 

\paragraph{Locomotion} The Locomotion datasets offer varying degrees of suboptimality, using mixtures of highly suboptimal trajectories (\texttt{-replay}) and optimal ones (\texttt{-expert}). Table~\ref{tab: gym locomotion results} shows BPR's Locomotion scores. In general, all methods recover near-expert performance on any \texttt{expert} datasets. BPR greatly outscores all older baselines as well as preference-based algorithms. ReBRAC is highly tuned for each dataset and BPR, for the most part, scores similarly except for \texttt{hc-m} (where ReBRAC scores higher), and \texttt{w-m} and \texttt{w-m-r}, where BPR outperforms ReBRAC by a substantial margin.

\begin{table*}
    \centering
    \begin{tabular}{l|rrr|rrr|rrr|r}
        \toprule
        Dataset & CQL & IQL & TD3+BC & ReBRAC & XQL & Diff-QL & PT & OPPO & DPPO & \textbf{BPR (ours)}
        \\
        \midrule
        \texttt{hc-m} & 44.0 & 47.4 & 48.3 & \textbf{65.6} & 48.3 & 51.1 & - & 43.4 & - & 53.7 $\pm$ 1.4
        \\
        \texttt{hp-m} & 58.5 & 66.3 & 59.3 & \textbf{102.0} & 74.2 & 90.5 & - & 86.3 & - & \underline{101.3 $\pm$ 1.1}
        \\
        \texttt{w-m} & 72.5 & 78.3 & 83.7 & 82.5 & 84.2 & 87.0 & - & 85.0 & - & \textbf{91.1 $\pm$ 3.7}
        \\
        \texttt{hc-m-r} & 45.5 & 42.2 & 44.6 & \textbf{51.0} & 45.2 & 47.8 & - & 39.8 & 40.8 & \underline{50.9 $\pm$ 0.6}
        \\
        \texttt{hp-m-r} & 95.0 & 94.7 & 60.9 & 98.1 & 100.7 & 101.3 & 84.5 & 88.9 & 73.2 & \textbf{102.0 $\pm$ 4.9}
        \\
        \texttt{w-m-r} & 77.2 & 73.9 & 81.8 & 77.3 & 82.2 & 95.5 & 71.3 & 71.7 & 50.9 & \textbf{97.4 $\pm$ 2.7}
        \\
        \texttt{hc-m-e} & 91.6 & 86.7 & 90.7 & 101.1 & 94.2 & 96.8 & - & 89.6 & 92.6 & \textbf{103.8 $\pm$ 4.3}
        \\
        \texttt{h-m-e} & 105.4 & 91.5 & 98.0 & 107.0 & \textbf{111.2 }& 111.1 & 69.0 & 108.0 & 107.2 & \underline{110.9 $\pm$ 5.2}
        \\
        \texttt{w-m-e} & 108.8 & 109.6 & 110.1 & 111.6 & \textbf{112.7} & 110.1 & 110.1 & 105.0 & 108.6 & 110.8 $\pm$ 0.2
    \end{tabular}
    \caption{Normalized scores on D4RL Gym Locomotion datasets. All scores are taken from their respective original papers. \texttt{hc}, \texttt{hp} and \texttt{w} refer to \texttt{halfcheetah}, \texttt{hopper} and \texttt{walker2d} environments, respectively. Methods are grouped by: older baselines, newer offline RL baselines, preference-based offline RL methods followed by BPR. For XQL, we use the per-dataset tuned variant's scores. We report SD for BPR and \textbf{bold} the top score and \underline{underline} BPR scores when within 1 SD of the best.}
    \label{tab: gym locomotion results}
\end{table*}

\paragraph{Antmaze} The Antmaze tasks are characterized by sparse reward schemes and suboptimal trajectories which necessitates off-policy evaluation (or IQL/XQL in-sample max estimation) to perform well. In the smaller mazes, BPR, ReBRAC and XQL perform similarly, though BPR is able to sustain high performance as the maze grows. Preference-based PT does not perform well in larger mazes. 

\begin{table*}
    \centering
    \begin{tabular}{l|rrr|rrr|r|r}
        \toprule
        Dataset & CQL & IQL & TD3+BC & ReBRAC & XQL & Diff-QL & PT & \textbf{BPR (ours)}
        \\
        \midrule
        \texttt{-umaze} & 74.0 & 87.5 & 78.6 & \textbf{97.8} & 93.8 & 93.4 & - & 95.6 $\pm$ 1.0
        \\
        \texttt{-umaze-d} & 84.0 & 62.2 & 71.4 & 88.3 & 82.0 & 66.2 & - & \textbf{89.1 $\pm$ 1.1}
        \\
        \texttt{-medium-p} & 61.2 & 71.2 & 10.6 & 84.0 & 76.0 & 76.6 & 70.1 & \textbf{86.7 $\pm$ 3.7}
        \\
        \texttt{-medium-d} & 53.7 & 70.0 & 3.0 & 76.3 & 73.6 & 78.6 & 65.3 & \textbf{82.9 $\pm$ 7.8}
        \\
        \texttt{-large-p} & 15.8 & 39.6 & 0.2 & 60.4 & 46.5 & 46.4 & 42.4 & \textbf{70.3 $\pm$ 8.3 }
        \\
        \texttt{-large-d} & 14.9 & 47.5 & 0.0 & 54.4 & 49.0 & 56.6 & 19.6 & \textbf{72.1 $\pm$ 5.1}
    \end{tabular}
    \caption{Normalized scores on D4RL Antmaze datasets. Methods are grouped by: older baselines, newer RM RL baselines, preference-based offline RL methods followed by BPR. For XQL, we use the per-dataset tuned variant's scores. We report SD for BPR and \textbf{bold} the top score and \underline{underline} BPR scores when within 1 SD of the best.}
    \label{tab: antmaze results}
\end{table*}

\subsection{V-D4RL}

Most offline RL algorithms typically limit their evaluation to proprioceptive state spaces. V-D4RL \citep{RN1000} is a benchmarking suite that evaluates offline RL algorithms in visual state spaces on continuous control tasks with mixtures of trajectories similar to those found in D4RL Locomotion and based on the DMC environments \citep{RN1001}. 

The V-D4RL paper provides scores for CQL, and behavioral cloning (BC) policies, as well as LOMPO \citep{RN1002} and a variant of DrQ \citep{RN1001} with a behavioral cloning constraint. LOMPO and DrQ are designed specifically to learn from visual state spaces. We also include results for ReBRAC, which is again tuned for each dataset. We use V-D4RL environments without distractors following \citet{RN842}. 

We present V-D4RL results in Table~\ref{tab: v-d4rl results}. Generally, BC outperforms CQL~-- the standard offline RL baseline. ReBRAC, with the help of tuning, is able to slightly outperform BC. BPR consistently outperforms the image-adapted LOMPO and DrQ+BC, trading blows with ReBRAC on \texttt{walker-walk} and \texttt{cheetah-run} datasets and keeps pace with BC on the more difficult \texttt{humanoid-walk} tasks. 

\begin{table*}
    \centering
    \begin{tabular}{l|r|rr|rr|r}
        \toprule
        Dataset & BC & CQL & ReBRAC & LOMPO & DrQ+BC & \textbf{BPR (ours)}
        \\
        \midrule
        \texttt{ww-mixed} & 16.5 $\pm$ 4.3 & 11.4 $\pm$ 12.4 & 41.6 $\pm$ 8.0 & 34.7 $\pm$ 19.7 & 28.7 $\pm$ 6.9 & \textbf{45.0 $\pm$ 11.2}
        \\
        \texttt{ww-medium} & 40.9 $\pm$ 3.1 & 14.8 $\pm$ 16.1 & \textbf{52.5 $\pm$ 3.2} & 43.9 $\pm$ 11.1 & 46.8 $\pm$ 2.3 & \underline{50.7 $\pm$ 4.1}
        \\
        \texttt{ww-medexp} & 47.7 $\pm$ 3.9 & 56.4 $\pm$ 38.4 & 92.7 $\pm$ 1.3 & 39.2 $\pm$ 19.5 & 86.4 $\pm$ 5.6 & \textbf{97.4 $\pm$ 1.9}
        \\
        \texttt{cr-mixed} & 25.0 $\pm$ 3.6 & 10.7 $\pm$ 12.8 & \textbf{46.8 $\pm$ 0.7} & 36.3 $\pm$ 15.6 & 44.8 $\pm$ 3.6 & \underline{45.0 $\pm$ 3.1}
        \\
        \texttt{cr-medium} & 51.6 $\pm$ 1.4 & 40.9 $\pm$ 5.1 & \textbf{58.3 $\pm$ 11.7} & 16.4 $\pm$ 18.3 & 50.6 $\pm$ 8.2 & 55.3 $\pm$ 1.2
        \\
        \texttt{cr-medexp} & 57.5 $\pm$ 6.3 & 20.9 $\pm$ 5.5 & 58.3 $\pm$ 11.7 & 11.9 $\pm$ 1.9 & 50.6 $\pm$ 8.2 & \textbf{62.7 $\pm$ 8.5}
        \\
        \texttt{hw-mixed} & \textbf{18.8 $\pm$ 4.2} & 0.1 $\pm$ 0.0 & 16.0 $\pm$ 2.7 & 0.2 $\pm$ 0.0 & 15.9 $\pm$ 3.8 & \underline{18.3 $\pm$ 1.9}
        \\
        \texttt{hw-medium} & \textbf{13.5 $\pm$ 4.1} & 0.1 $\pm$ 0.0 & 9.0 $\pm$ 2.3 & 0.1 $\pm$ 0.0 & 6.2 $\pm$ 2.4 & 9.0 $\pm$ 0.8
        \\
        \texttt{hw-medexp} & \textbf{17.2 $\pm$ 4.7} & 0.1 $\pm$ 0.0 & 7.8 $\pm$ 2.4 & 0.2 $\pm$ 0.0 & 7.0 $\pm$ 2.3 & \underline{13.3 $\pm$ 4.4}
    \end{tabular}
    \caption{Normalized scores on V-D4RL tasks. \texttt{ww}, \texttt{cr} and \texttt{hw} refer to \texttt{walker-walk}, \texttt{cheetah-run} and \texttt{humanoid-walk} environments, respectively. Methods are grouped by: BC, offline RL baselines, RL algorithms adapted for visual state spaces followed by BPR. We report 1 SD for all methods and \textbf{bold} the top score and \underline{underline} BPR scores when within 1 SD of the best.}
    \label{tab: v-d4rl results}
\end{table*}

\subsection{Onestep Experiments}
\label{subsec: onestep experiments}

Off-policy evaluation can lead to querying and backing up of overestimated OOD actions that the policy can exploit, leading to instability. Onestep value functions are highly stable due to their on-policy nature \citep{RN882} and recent work by \citet{RN990} shows equivalence between Onestep values and CQL-style critic regularization.

We evaluate how well BPR with a Onestep value function performs compared to the original Onestep RL (O-RL) algorithm \citep{RN882}. We also include Locomotion results from CFPI \citep{RN876}, which uses a first-order Taylor approximation as a linear approximation of the Q function, and trains a Onestep value function using distributional critics \citep{RN1003,RN1004}.

We report results on non-expert Locomotion datasets and the \texttt{medium} and \texttt{large} Antmaze datasets in Table~\ref{tab: onestep results}. Both Onestep RL and CFPI perform similarly on Locomotion tasks. BPR matches their performance on two tasks and outperforms both by a large margin on four out of six Locomotion tasks.

Onestep RL performs poorly on the \texttt{medium} and \texttt{large} Antmaze tasks. In contrast, BPR is able to make significant progress in all these sparse reward tasks, falling slightly short of off-policy CQL (see Table~\ref{tab: antmaze results}). 

\paragraph{Suboptimality in D4RL} The similarity in performance between Onestep BPR and off-policy BPR in Locomotion tasks suggests that trajectories in these datasets may not be as suboptimal as originally thought \citep{RN731}. This explains the recent saturation in performance on Locomotion \citep{RN842}. Antmaze, while challengingly suboptimal, may be a poor evaluator of generalization \citep{RN936}. The performance of Onestep BPR indicates that this may be a pragmatic variant to select for application due to its improved stability. 

\paragraph{More Expressive Onestep Value Functions} O-RL uses a single Q function and samples actions to estimate state-value to compute advantage. CFPI trains two distributional critics and ses the min-clipped value estimate during bootstrapping. Onestep BPR trains two regular, min-clipped critics. Diversity can collapse in ensembles with shared targets. We investigate whether diversity at the cost of pessimism can improve performance; we experiment with Onestep, independent 4-critic ensembles to estimate the Q value lower confidence bound \citep{RN771}:
\begin{align}
    Q_{\text{LCB}} (s, a) = \mathbb{E}^{\text{ens}} \left [Q_i (s, a) \right] - \omega \mathbb{V}^{\text{ens}} \left[ Q_i (s, a) \right],
\end{align}

\noindent where $\mathbb{E}^{\text{ens}}$ and $\mathbb{V}^{\text{ens}}$ indicate mean and variance over the ensemble of Q functions and $\omega$ is a parameter that controls the degree of pessimism. We use $\omega = 2.0$ in all experiments.

Compared to Onestep BPR, Ensemble BPR sees performance improvements of \textbf{at least 10 points on each dataset} on the \texttt{medium} and \texttt{large} Antmaze datasets. Detailed per-dataset scores and implementation information can be found in the Appendix. 

\begin{table}
    \centering
    \begin{tabular}{lrrr}
        \toprule
        Dataset & O-RL & CFPI & \textbf{Onestep BPR}
        \\
        \midrule
        \texttt{hc-m} & \textbf{55.6} & 51.1 & 52.0 $\pm$ 0.8
        \\
        \texttt{hp-m} & 83.3 & 86.8 & \textbf{96.4 $\pm$ 0.4}
        \\
        \texttt{w-m} & 85.6 & 88.3 & \textbf{89.7 $\pm$ 1.3}
        \\
        \texttt{hc-m-r} & 41.4 & 44.5 & \textbf{51.0 $\pm$ 0.4}
        \\
        \texttt{h-m-r} & 71.0 & 93.6 & \textbf{99.1 $\pm$ 2.3}
        \\
        \texttt{w-m-r} & 71.6 & 78.2 & \textbf{92.0 $\pm$ 0.8}
        \\
        \texttt{amaze-m-p} & 0.3 & - & \textbf{52.7 $\pm$ 10.3}
        \\
        \texttt{amaze-m-d} & 0.0 & - & \textbf{40.0 $\pm$ 7.8}
        \\
        \texttt{amaze-l-p} & 0.0 & - & \textbf{10.4 $\pm$ 2.9}
        \\
        \texttt{amaze-l-d} & 0.0 & - & \textbf{12.7 $\pm$ 1.6}
    \end{tabular}
    \caption{Scores for Onestep BPR with Onestep RL and Onestep CFPI. We evaluate on non-expert Locomotion and \texttt{medium} and \texttt{large} Antmaze (\texttt{amaze}) datasets. The authors of CFPI do not report Onestep results for Antmaze. We report 1 SD for BPR and \textbf{bold} the top score and \underline{underline} BPR scores when within 1 SD of the best.}
    \label{tab: onestep results}
\end{table}

\subsection{Ablations}

Recall that $\lambda$ controls the tradeoff between maximizing
behavioral consistency and fitting the Q function in Equation~\ref{eq: KL constrained problem}. We examine sensitivity to $\lambda$ for off-policy BPR in a series of ablation experiments in the D4RL Locomotion tasks. 

Sensitivity to $\lambda$ varies between datasets, with little performance variation on \texttt{halfcheetah-medium} and \texttt{halfcheetah-medium-replay}. In other datasets, using $\lambda = 0.5$ or $\lambda = 2.0$ sees performance decline. Our choice of $\lambda = 1.0$ generalizes well over all datasets and usually outperforms $\lambda = 1.5$. We provide detailed ablation results in the Appendix.

\section{Discussion}

\paragraph{Performance} Our key contribution in this work is the development of a policy objective that reduces policy improvement to a regression problem. Off-policy BPR results in D4RL Locomotion datasets are on par with current SOTA and BPR \textbf{outperforms RL baselines in 5 out of 6 Antmaze datasets} and \textbf{6 out of 9 V-D4RL datasets}. Our Onestep experiments show that Onestep BPR outperforms Onestep RL in \textbf{9 out of 10 tasks} and CFPI in all Locomotion tasks. BPR requires \textbf{minimal tuning} to achieve high performance~-- \textbf{all our results are produced using $\mathbf{\lambda = 1.0}$}.

\paragraph{Density Estimation} Employing estimates of the behavior policy is common in many offline RL algorithms. Most prior works use explicit density estimates using Gaussian policies, mixture density networks \citep{RN460} or VAEs \citep{RN521}. If the modality of the behavior policy is known, the first two methods can be used in BPR. VAEs are unsuitable as density estimation requires sampling.

The function $f(\cdot, \cdot)$ does not need to be a density estimate. Another natural choice for $f(\cdot, \cdot)$ is a discriminator \citep{RN458} that replaces a density estimate with an adversarial critic trained concurrently. This offers more choice of the exact \textit{f}-divergence to minimize at the cost of increased training instability \citep{RN1006}.

\paragraph{Critic Ensembles} Our ensemble experiments imply that Onestep-trained policies might perform better than prior work reports. The \textit{optimistic pessimism} of $Q_{\text{LCB}}$ ensembles could enable algorithms to learn better policies while still enjoying the stability of on-policy evaluation. 

\paragraph{Limitations} EBMs can be difficult and computationally expensive to train. As a consequence of the Manifold hypothesis, they may also generalize poorly \citep{RN573}, though all models capable of multimodal learning suffer from their own slew of problems \citep{RN884}. Advancements in methodology have improved the stability of training and quality of models \citep{RN1007}. Both prior work \citep{RN885} and the results of our experiments suggest that EBMs are well-suited for offline RL.

\section{Conclusion}

In this paper, we introduce Behavior Preference Regression (BPR). Our method formulates a reframed, paired-sample policy objective that directly trains a policy likelihood to be behaviorally consistent and maximize reward, using least-squares regression. Though our method is motivated by finetuning approaches in language models, it is extensible to offline RL. We validate our algorithm on datasets with a variety of task types and reward schemes that offer both proprioceptive and image-based state spaces. BPR consistently outperforms prior RM-based approaches and preference-based ones by a substantial margin.

Additional experiments evaluating Onestep BPR demonstrate that our algorithm can learn policies that outperform previous Onestep methods. Furthermore, with more expressive Onestep value functions, BPR makes headway on the challenging Antmaze tasks that typically demand off-policy evaluation. 

Future work should further review the viability of Onestep ensembles and look to adapt paired completion approaches for offline continuous control.

\bibliography{mybib}

\begin{thebibliography}{62}
\providecommand{\natexlab}[1]{#1}

\bibitem[{Akrour, Schoenauer, and Sebag(2011)}]{RN983}
Akrour, R.; Schoenauer, M.; and Sebag, M. 2011.
\newblock Preference-based policy learning.
\newblock In \emph{Machine Learning and Knowledge Discovery in Databases: European Conference, ECML PKDD 2011, Athens, Greece, September 5-9, 2011. Proceedings, Part I 11}, 12--27. Springer.
\newblock ISBN 3642237797.

\bibitem[{An et~al.(2023)An, Lee, Zuo, Kosaka, Kim, and Song}]{RN1115}
An, G.; Lee, J.; Zuo, X.; Kosaka, N.; Kim, K.-M.; and Song, H.~O. 2023.
\newblock Direct preference-based policy optimization without reward modeling.
\newblock \emph{Advances in Neural Information Processing Systems}, 36: 70247--70266.

\bibitem[{An et~al.(2021)An, Moon, Kim, and Song}]{RN770}
An, G.; Moon, S.; Kim, J.-H.; and Song, H.~O. 2021.
\newblock Uncertainty-based offline reinforcement learning with diversified {Q}-ensemble.
\newblock \emph{Advances in Neural Information Processing Systems}, 34: 7436--7447.

\bibitem[{Bengio, Courville, and Vincent(2013)}]{RN573}
Bengio, Y.; Courville, A.; and Vincent, P. 2013.
\newblock Representation learning: A review and new perspectives.
\newblock \emph{IEEE transactions on pattern analysis and machine intelligence}, 35(8): 1798--1828.

\bibitem[{Bishop(1994)}]{RN460}
Bishop, C.~M. 1994.
\newblock Mixture density networks.

\bibitem[{Bradley and Terry(1952)}]{RN937}
Bradley, R.~A.; and Terry, M.~E. 1952.
\newblock Rank analysis of incomplete block designs: I. The method of paired comparisons.
\newblock \emph{Biometrika}, 39(3/4): 324--345.

\bibitem[{Brandfonbrener et~al.(2021)Brandfonbrener, Whitney, Ranganath, and Bruna}]{RN882}
Brandfonbrener, D.; Whitney, W.; Ranganath, R.; and Bruna, J. 2021.
\newblock Offline {RL} without off-policy evaluation.
\newblock \emph{Advances in Neural Information Processing Systems}, 34: 4933--4946.

\bibitem[{Cheng et~al.(2011)Cheng, Fürnkranz, Hüllermeier, and Park}]{RN984}
Cheng, W.; Fürnkranz, J.; Hüllermeier, E.; and Park, S.-H. 2011.
\newblock Preference-based policy iteration: Leveraging preference learning for reinforcement learning.
\newblock In \emph{Machine Learning and Knowledge Discovery in Databases: European Conference, ECML PKDD 2011, Athens, Greece, September 5-9, 2011. Proceedings, Part I 11}, 312--327. Springer.
\newblock ISBN 3642237797.

\bibitem[{Christiano et~al.(2017)Christiano, Leike, Brown, Martic, Legg, and Amodei}]{RN965}
Christiano, P.~F.; Leike, J.; Brown, T.; Martic, M.; Legg, S.; and Amodei, D. 2017.
\newblock Deep reinforcement learning from human preferences.
\newblock \emph{Advances in Neural Information Processing Systems}, 30.

\bibitem[{Dabney et~al.(2018{\natexlab{a}})Dabney, Ostrovski, Silver, and Munos}]{RN1003}
Dabney, W.; Ostrovski, G.; Silver, D.; and Munos, R. 2018{\natexlab{a}}.
\newblock Implicit quantile networks for distributional reinforcement learning.
\newblock In \emph{International Conference on Machine Learning}, 1096--1105. PMLR.
\newblock ISBN 2640-3498.

\bibitem[{Dabney et~al.(2018{\natexlab{b}})Dabney, Rowland, Bellemare, and Munos}]{RN1004}
Dabney, W.; Rowland, M.; Bellemare, M.; and Munos, R. 2018{\natexlab{b}}.
\newblock Distributional reinforcement learning with quantile regression.
\newblock In \emph{Proceedings of the {AAAI} conference on artificial intelligence}, volume~32.
\newblock ISBN 2374-3468.

\bibitem[{Du and Mordatch(2019)}]{RN1007}
Du, Y.; and Mordatch, I. 2019.
\newblock Implicit generation and modeling with energy based models.
\newblock \emph{Advances in Neural Information Processing Systems}, 32.

\bibitem[{Eysenbach et~al.(2023)Eysenbach, Geist, Levine, and Salakhutdinov}]{RN990}
Eysenbach, B.; Geist, M.; Levine, S.; and Salakhutdinov, R. 2023.
\newblock A connection between one-step RL and critic regularization in reinforcement learning.
\newblock In \emph{International Conference on Machine Learning}, 9485--9507. PMLR.
\newblock ISBN 2640-3498.

\bibitem[{Florence et~al.(2022)Florence, Lynch, Zeng, Ramirez, Wahid, Downs, Wong, Lee, Mordatch, and Tompson}]{RN885}
Florence, P.; Lynch, C.; Zeng, A.; Ramirez, O.~A.; Wahid, A.; Downs, L.; Wong, A.; Lee, J.; Mordatch, I.; and Tompson, J. 2022.
\newblock Implicit behavioral cloning.
\newblock In \emph{Conference on Robot Learning}, 158--168. PMLR.
\newblock ISBN 2640-3498.

\bibitem[{Fu et~al.(2020)Fu, Kumar, Nachum, Tucker, and Levine}]{RN731}
Fu, J.; Kumar, A.; Nachum, O.; Tucker, G.; and Levine, S. 2020.
\newblock {D4RL}: Datasets for deep data-driven reinforcement learning.
\newblock \emph{arXiv preprint arXiv:2004.07219}.

\bibitem[{Fu, Wu, and Boulet(2022)}]{RN813}
Fu, Y.; Wu, D.; and Boulet, B. 2022.
\newblock A closer look at offline {RL} agents.
\newblock \emph{Advances in Neural Information Processing Systems}, 35: 8591--8604.

\bibitem[{Fujimoto and Gu(2021)}]{RN719}
Fujimoto, S.; and Gu, S.~S. 2021.
\newblock A minimalist approach to offline reinforcement learning.
\newblock \emph{Advances in Neural Information Processing Systems}, 34: 20132--20145.

\bibitem[{Fujimoto, Hoof, and Meger(2018)}]{RN714}
Fujimoto, S.; Hoof, H.; and Meger, D. 2018.
\newblock Addressing function approximation error in actor-critic methods.
\newblock In \emph{International Conference on Machine Learning}, 1587--1596. PMLR.
\newblock ISBN 2640-3498.

\bibitem[{Fujimoto, Meger, and Precup(2019)}]{RN684}
Fujimoto, S.; Meger, D.; and Precup, D. 2019.
\newblock Off-policy deep reinforcement learning without exploration.
\newblock In \emph{International Conference on Machine Learning}, 2052--2062. PMLR.
\newblock ISBN 2640-3498.

\bibitem[{Gao et~al.(2024)Gao, Chang, Zhan, Oertell, Swamy, Brantley, Joachims, Bagnell, Lee, and Sun}]{RN942}
Gao, Z.; Chang, J.~D.; Zhan, W.; Oertell, O.; Swamy, G.; Brantley, K.; Joachims, T.; Bagnell, J.~A.; Lee, J.~D.; and Sun, W. 2024.
\newblock {REBEL}: Reinforcement Learning via Regressing Relative Rewards.
\newblock \emph{arXiv preprint arXiv:2404.16767}.

\bibitem[{Garg et~al.(2023)Garg, Hejna, Geist, and Ermon}]{RN954}
Garg, D.; Hejna, J.; Geist, M.; and Ermon, S. 2023.
\newblock Extreme {Q}-learning: Maxent {RL} without entropy.
\newblock \emph{arXiv preprint arXiv:2301.02328}.

\bibitem[{Geng et~al.(2022)Geng, Li, Gupta, Kumar, and Levine}]{RN979}
Geng, X.; Li, K.; Gupta, A.; Kumar, A.; and Levine, S. 2022.
\newblock Effective offline {RL} needs going beyond pessimism: Representations and distributional shift.
\newblock In \emph{Decision Awareness in Reinforcement Learning Workshop at ICML 2022}.

\bibitem[{Ghasemipour, Gu, and Nachum(2022)}]{RN771}
Ghasemipour, K.; Gu, S.~S.; and Nachum, O. 2022.
\newblock Why so pessimistic? estimating uncertainties for offline {RL} through ensembles, and why their independence matters.
\newblock \emph{Advances in Neural Information Processing Systems}, 35: 18267--18281.

\bibitem[{Goodfellow, Bengio, and Courville(2016)}]{RN884}
Goodfellow, I.; Bengio, Y.; and Courville, A. 2016.
\newblock \emph{Deep learning}.
\newblock MIT press.
\newblock ISBN 0262337371.

\bibitem[{Goodfellow et~al.(2014)Goodfellow, Pouget-Abadie, Mirza, Xu, Warde-Farley, Ozair, Courville, and Bengio}]{RN458}
Goodfellow, I.; Pouget-Abadie, J.; Mirza, M.; Xu, B.; Warde-Farley, D.; Ozair, S.; Courville, A.; and Bengio, Y. 2014.
\newblock Generative adversarial nets.
\newblock \emph{Advances in Neural Information Processing Systems}, 27.

\bibitem[{Grünwald and Dawid(2004)}]{RN987}
Grünwald, P.~D.; and Dawid, A.~P. 2004.
\newblock Game theory, maximum entropy, minimum discrepancy and robust Bayesian decision theory.

\bibitem[{Gulcehre et~al.(2020)Gulcehre, Colmenarejo, Sygnowski, Paine, Zolna, Chen, Hoffman, Pascanu, and de~Freitas}]{RN841}
Gulcehre, C.; Colmenarejo, S.~G.; Sygnowski, J.; Paine, T.; Zolna, K.; Chen, Y.; Hoffman, M.; Pascanu, R.; and de~Freitas, N. 2020.
\newblock Addressing Extrapolation Error in Deep Offline Reinforcement Learning.

\bibitem[{Haarnoja et~al.(2018)Haarnoja, Zhou, Abbeel, and Levine}]{RN802}
Haarnoja, T.; Zhou, A.; Abbeel, P.; and Levine, S. 2018.
\newblock Soft actor-critic: Off-policy maximum entropy deep reinforcement learning with a stochastic actor.
\newblock In \emph{International Conference on Machine Learning}, 1861--1870. PMLR.
\newblock ISBN 2640-3498.

\bibitem[{Hejna et~al.(2023)Hejna, Rafailov, Sikchi, Finn, Niekum, Knox, and Sadigh}]{RN938}
Hejna, J.; Rafailov, R.; Sikchi, H.; Finn, C.; Niekum, S.; Knox, W.~B.; and Sadigh, D. 2023.
\newblock Contrastive prefence learning: Learning from human feedback without {RL}.
\newblock \emph{arXiv preprint arXiv:2310.13639}.

\bibitem[{Ho, Jain, and Abbeel(2020)}]{RN623}
Ho, J.; Jain, A.; and Abbeel, P. 2020.
\newblock Denoising diffusion probabilistic models.
\newblock \emph{Advances in Neural Information Processing Systems}, 33: 6840--6851.

\bibitem[{Jin, Yang, and Wang(2021)}]{RN782}
Jin, Y.; Yang, Z.; and Wang, Z. 2021.
\newblock Is pessimism provably efficient for offline {RL}?
\newblock In \emph{International Conference on Machine Learning}, 5084--5096. PMLR.
\newblock ISBN 2640-3498.

\bibitem[{Jolicoeur-Martineau(2020)}]{RN1006}
Jolicoeur-Martineau, A. 2020.
\newblock On relativistic f-divergences.
\newblock In \emph{International Conference on Machine Learning}, 4931--4939. PMLR.
\newblock ISBN 2640-3498.

\bibitem[{Kang et~al.(2023)Kang, Shi, Liu, He, and Wang}]{RN997}
Kang, Y.; Shi, D.; Liu, J.; He, L.; and Wang, D. 2023.
\newblock Beyond reward: Offline preference-guided policy optimization.
\newblock \emph{arXiv preprint arXiv:2305.16217}.

\bibitem[{Kaufmann et~al.(2023)Kaufmann, Weng, Bengs, and Hüllermeier}]{RN985}
Kaufmann, T.; Weng, P.; Bengs, V.; and Hüllermeier, E. 2023.
\newblock A survey of reinforcement learning from human feedback.
\newblock \emph{arXiv preprint arXiv:2312.14925}.

\bibitem[{Kim et~al.(2023)Kim, Park, Shin, Lee, Abbeel, and Lee}]{RN939}
Kim, C.; Park, J.; Shin, J.; Lee, H.; Abbeel, P.; and Lee, K. 2023.
\newblock Preference transformer: Modeling human preferences using transformers for {RL}.
\newblock \emph{arXiv preprint arXiv:2303.00957}.

\bibitem[{Kingma and Welling(2013)}]{RN521}
Kingma, D.~P.; and Welling, M. 2013.
\newblock Auto-encoding variational bayes.
\newblock \emph{arXiv preprint arXiv:1312.6114}.

\bibitem[{Knox et~al.(2022)Knox, Hatgis-Kessell, Booth, Niekum, Stone, and Allievi}]{RN992}
Knox, W.~B.; Hatgis-Kessell, S.; Booth, S.; Niekum, S.; Stone, P.; and Allievi, A. 2022.
\newblock Models of human preference for learning reward functions.
\newblock \emph{arXiv preprint arXiv:2206.02231}.

\bibitem[{Kostrikov et~al.(2021)Kostrikov, Fergus, Tompson, and Nachum}]{RN918}
Kostrikov, I.; Fergus, R.; Tompson, J.; and Nachum, O. 2021.
\newblock Offline reinforcement learning with {F}isher divergence critic regularization.
\newblock In \emph{International Conference on Machine Learning}, 5774--5783. PMLR.
\newblock ISBN 2640-3498.

\bibitem[{Kostrikov, Nair, and Levine(2021)}]{RN711}
Kostrikov, I.; Nair, A.; and Levine, S. 2021.
\newblock Offline reinforcement learning with implicit {Q}-learning.
\newblock \emph{arXiv preprint arXiv:2110.06169}.

\bibitem[{Kumar et~al.(2020)Kumar, Zhou, Tucker, and Levine}]{RN698}
Kumar, A.; Zhou, A.; Tucker, G.; and Levine, S. 2020.
\newblock Conservative {Q}-learning for offline reinforcement learning.
\newblock \emph{Advances in Neural Information Processing Systems}, 33: 1179--1191.

\bibitem[{Lange, Gabel, and Riedmiller(2012)}]{RN695}
Lange, S.; Gabel, T.; and Riedmiller, M. 2012.
\newblock \emph{Batch reinforcement learning}, 45--73.
\newblock Springer.

\bibitem[{Li et~al.(2023)Li, Zhang, Yin, Bai, Wang, and Wang}]{RN876}
Li, J.; Zhang, E.; Yin, M.; Bai, Q.; Wang, Y.-X.; and Wang, W.~Y. 2023.
\newblock Offline reinforcement learning with closed-form policy improvement operators.
\newblock In \emph{International Conference on Machine Learning}, 20485--20528. PMLR.
\newblock ISBN 2640-3498.

\bibitem[{Lu et~al.(2022)Lu, Ball, Rudner, Parker-Holder, Osborne, and Teh}]{RN1000}
Lu, C.; Ball, P.~J.; Rudner, T.~G.; Parker-Holder, J.; Osborne, M.~A.; and Teh, Y.~W. 2022.
\newblock Challenges and opportunities in offline reinforcement learning from visual observations.
\newblock \emph{arXiv preprint arXiv:2206.04779}.

\bibitem[{Miyato et~al.(2018)Miyato, Kataoka, Koyama, and Yoshida}]{RN888}
Miyato, T.; Kataoka, T.; Koyama, M.; and Yoshida, Y. 2018.
\newblock Spectral normalization for generative adversarial networks.
\newblock \emph{arXiv preprint arXiv:1802.05957}.

\bibitem[{Nair et~al.(2020)Nair, Gupta, Dalal, and Levine}]{RN706}
Nair, A.; Gupta, A.; Dalal, M.; and Levine, S. 2020.
\newblock {AWAC}: Accelerating online reinforcement learning with offline datasets.
\newblock \emph{arXiv preprint arXiv:2006.09359}.

\bibitem[{Peng et~al.(2019)Peng, Kumar, Zhang, and Levine}]{RN705}
Peng, X.~B.; Kumar, A.; Zhang, G.; and Levine, S. 2019.
\newblock Advantage-weighted regression: Simple and scalable off-policy reinforcement learning.
\newblock \emph{arXiv preprint arXiv:1910.00177}.

\bibitem[{Peters, Mulling, and Altun(2010)}]{RN993}
Peters, J.; Mulling, K.; and Altun, Y. 2010.
\newblock Relative entropy policy search.
\newblock In \emph{Proceedings of the {AAAI} Conference on Artificial Intelligence}, volume~24, 1607--1612.
\newblock ISBN 2374-3468.

\bibitem[{Rafailov et~al.(2024)Rafailov, Sharma, Mitchell, Manning, Ermon, and Finn}]{RN936}
Rafailov, R.; Sharma, A.; Mitchell, E.; Manning, C.~D.; Ermon, S.; and Finn, C. 2024.
\newblock Direct preference optimization: Your language model is secretly a reward model.
\newblock \emph{Advances in Neural Information Processing Systems}, 36.

\bibitem[{Rafailov et~al.(2021)Rafailov, Yu, Rajeswaran, and Finn}]{RN1002}
Rafailov, R.; Yu, T.; Rajeswaran, A.; and Finn, C. 2021.
\newblock Offline reinforcement learning from images with latent space models.
\newblock In \emph{Learning for dynamics and control}, 1154--1168. PMLR.
\newblock ISBN 2640-3498.

\bibitem[{Razzaghi et~al.(2022)Razzaghi, Tabrizian, Guo, Chen, Taye, Thompson, Bregeon, Baheri, and Wei}]{RN964}
Razzaghi, P.; Tabrizian, A.; Guo, W.; Chen, S.; Taye, A.; Thompson, E.; Bregeon, A.; Baheri, A.; and Wei, P. 2022.
\newblock A survey on reinforcement learning in aviation applications.
\newblock \emph{arXiv preprint arXiv:2211.02147}.

\bibitem[{Schulman et~al.(2017)Schulman, Wolski, Dhariwal, Radford, and Klimov}]{RN982}
Schulman, J.; Wolski, F.; Dhariwal, P.; Radford, A.; and Klimov, O. 2017.
\newblock Proximal policy optimization algorithms.
\newblock \emph{arXiv preprint arXiv:1707.06347}.

\bibitem[{Shalev-Shwartz, Shamir, and Shammah(2017)}]{RN967}
Shalev-Shwartz, S.; Shamir, O.; and Shammah, S. 2017.
\newblock Failures of gradient-based deep learning.
\newblock In \emph{International Conference on Machine Learning}, 3067--3075. PMLR.
\newblock ISBN 2640-3498.

\bibitem[{Sutton and Barto(2018)}]{RN679}
Sutton, R.~S.; and Barto, A.~G. 2018.
\newblock \emph{Reinforcement learning: An introduction}.
\newblock MIT press.
\newblock ISBN 0262352702.

\bibitem[{Swamy et~al.(2024)Swamy, Dann, Kidambi, Wu, and Agarwal}]{RN989}
Swamy, G.; Dann, C.; Kidambi, R.; Wu, Z.~S.; and Agarwal, A. 2024.
\newblock A minimaximalist approach to reinforcement learning from human feedback.
\newblock \emph{arXiv preprint arXiv:2401.04056}.

\bibitem[{Tarasov et~al.(2023)Tarasov, Kurenkov, Nikulin, and Kolesnikov}]{RN842}
Tarasov, D.; Kurenkov, V.; Nikulin, A.; and Kolesnikov, S. 2023.
\newblock Revisiting the Minimalist Approach to Offline Reinforcement Learning.
\newblock \emph{arXiv preprint arXiv:2305.09836}.

\bibitem[{Vieillard et~al.(2021)Vieillard, Andrychowicz, Raichuk, Pietquin, and Geist}]{RN956}
Vieillard, N.; Andrychowicz, M.; Raichuk, A.; Pietquin, O.; and Geist, M. 2021.
\newblock Implicitly regularized {RL} with implicit {Q}-values.
\newblock \emph{arXiv preprint arXiv:2108.07041}.

\bibitem[{Wang, Hunt, and Zhou(2022)}]{RN999}
Wang, Z.; Hunt, J.~J.; and Zhou, M. 2022.
\newblock Diffusion policies as an expressive policy class for offline reinforcement learning.
\newblock \emph{arXiv preprint arXiv:2208.06193}.

\bibitem[{Wang et~al.(2020)Wang, Novikov, Zolna, Merel, Springenberg, Reed, Shahriari, Siegel, Gulcehre, and Heess}]{RN708}
Wang, Z.; Novikov, A.; Zolna, K.; Merel, J.~S.; Springenberg, J.~T.; Reed, S.~E.; Shahriari, B.; Siegel, N.; Gulcehre, C.; and Heess, N. 2020.
\newblock Critic regularized regression.
\newblock \emph{Advances in Neural Information Processing Systems}, 33: 7768--7778.

\bibitem[{Wu, Tucker, and Nachum(2019)}]{RN699}
Wu, Y.; Tucker, G.; and Nachum, O. 2019.
\newblock Behavior regularized offline reinforcement learning.
\newblock \emph{arXiv preprint arXiv:1911.11361}.

\bibitem[{Yarats et~al.(2021)Yarats, Fergus, Lazaric, and Pinto}]{RN1001}
Yarats, D.; Fergus, R.; Lazaric, A.; and Pinto, L. 2021.
\newblock Mastering visual continuous control: Improved data-augmented reinforcement learning.
\newblock \emph{arXiv preprint arXiv:2107.09645}.

\bibitem[{Zhuang et~al.(2023)Zhuang, Lei, Liu, Wang, and Guo}]{RN943}
Zhuang, Z.; Lei, K.; Liu, J.; Wang, D.; and Guo, Y. 2023.
\newblock Behavior proximal policy optimization.
\newblock \emph{arXiv preprint arXiv:2302.11312}.

\bibitem[{Ziebart et~al.(2008)Ziebart, Maas, Bagnell, and Dey}]{RN935}
Ziebart, B.~D.; Maas, A.~L.; Bagnell, J.~A.; and Dey, A.~K. 2008.
\newblock Maximum entropy inverse reinforcement learning.
\newblock In \emph{{AAAI}}, volume~8, 1433--1438. Chicago, IL, USA.

\end{thebibliography}

\end{document}


\onecolumn

\section{Proofs}

We have the implicit Q function $\tilde{Q}(s, \cdot) = Q(s, \cdot) + \frac{1}{\lambda} \log \pi_{\beta} (\cdot | s)$. 

\subsection{Proposition 1}

We begin with the following Lemma for the difference in return between two policies:
\begin{lemma}
    \label{lem: performance diff}
    Given two policies $\pi_1$ and $\pi_2$, we can show:
    \begin{align}
        \eta(\pi_1) - \eta(\pi_2) = \int_{s \in \mathcal{S}} \rho_{\pi_1}(s) (\tilde{Q}_{\pi_{2}} (s, \pi_{1} (s)) - V_{\pi_{2}} (s))\, ds
    \end{align}

\begin{proof}
    From \citet{RN1008} we know: 
    \begin{align}
        \eta({\pi_{1}}) &= \eta(\pi_{2}) + \mathbb{E}_{\tau \sim \rho_{\pi_{1}}} \left[ \sum_{t=0}^{\infty} \gamma^t \tilde{Q}_{\pi_{2}} (s_t, a_t) - V_{\pi_{2}} (s_t) \right]
        \\
        &= \eta(\pi_{2}) + \sum_{t=0}^{\infty} \int_{s \in \mathcal{S}} P(s_t = s | \pi_{1}) \gamma^t (\tilde{Q}_{\pi_{2}} (s_t, \pi_{1} (s) - V_{\pi_{2}} (s))\, ds
        \\
        &= \eta (\pi_{2}) + \int_{s \in \mathcal{S}} \sum_{t=0}^{\infty} \gamma^t P(s_t = s | \pi_{1}) (\tilde{Q}_{\pi_{2}} (s_t, \pi_{1} (s) - V_{\pi_{2}} (s))\, ds
        \\
        &= \eta (\pi_{2}) + \int_{s \in \mathcal{S}} \sum_{t=0}^{\infty} \rho_{\pi_{1}}(s) (\tilde{Q}_{\pi_{2}} (s_t, \pi_{1} (s) - V_{\pi_{2}} (s))\, ds.
    \end{align}
    Which proves Lemma~\ref{lem: performance diff}.
\end{proof}

\end{lemma}

Using Lemma~\ref{lem: performance diff} we can write:
\begin{align}
    \eta (\tilde{\pi}) - \eta (\pi^*) = \int_{s \in \mathcal{S}} \rho_{\tilde{\pi}} (s) (\tilde{Q}^* (s, \tilde{\pi} (s)) - V^* (s))\, ds,
\end{align}

\noindent and
\begin{align}
    \eta (\pi_{\beta}) - \eta (\pi^*) = \int_{s \in \mathcal{S}} \rho_{\pi_{\beta}} (s) (\tilde{Q}^* (s, \pi_{\beta} (s)) - V^* (s))\, ds,
\end{align}

\noindent where $\tilde{Q}^* = \tilde{Q}_{\pi^*} $and $V^* = V_{\pi^*}$. 

We compute the difference:
\begin{align}
    \eta (\tilde{\pi}) - \eta (\tilde{\pi_{\beta}}) &= \int_{s \in \mathcal{S}} \rho_{\tilde{\pi}} (s) (\tilde{Q}^* (s, \tilde{\pi} (s)) - V^* (s)) - \rho_{\pi_{\beta}} (s) (\tilde{Q}^* (s, \pi_{\beta} (s)) - V^* (s))\, ds,
\end{align}

\noindent and using $\rho_{\tilde{\pi}} \approx \rho_{\pi_{\beta}}$ we have:
\begin{align}
    \label{eq: q diff perfect}
    \eta (\tilde{\pi}) - \eta (\tilde{\pi_{\beta}}) \approx \int_{s \in \mathcal{S}} \rho_{\pi_{\beta}}(s) (\tilde{Q}^* (s, \tilde{\pi} (s)) - \tilde{Q}^* (s, \pi_{\beta} (s)))\, ds.
\end{align}

Then, $\forall s \in \mathcal{S}$, using Assumption~1 and $\rho_{\pi_{\beta}} \geq 0$ it follows that:
\begin{align}
    \eta (\tilde{\pi}) - \eta (\tilde{\pi_{\beta}}) &\approx \int_{s \in \mathcal{S}} \rho_{\pi_{\beta}}(s) (\tilde{Q}^* (s, \tilde{\pi} (s)) - \tilde{Q}^* (s, \pi_{\beta} (s)))\, ds \geq 0,
\end{align}

\noindent and for the empirical expectation:
\begin{align}
    \quad &= \mathbb{E}_{s \sim \mathcal{D}} \left[ \tilde{Q}^* (s, \tilde{\pi} (s)) - \tilde{Q}^* (s, \pi_{\beta} (s)) \right] \geq 0,
\end{align}

\noindent which completes the proof.

\subsection{Proposition 2}

We first state the following Lemma for Total Variation Distance (TVD):
\begin{lemma}
    \label{lem: tv distance}
    Let $u$ and $v$ be two probability distributions on $\mathcal{X}$, then from \citet{RN1009}:
    \begin{align}
        \max_{x \in \mathcal{X}} \lvert u(x)  - v(x) \rvert = \frac{1}{2} \sum_{x\in \mathcal{X}} \lvert u(x) - v(x) \rvert.
    \end{align}
    \begin{proof}
        See \citet{RN1009}, Proposition~5.2 and related proof.
    \end{proof}
\end{lemma}

In practice, when using approximations of the Q function and $\pi_{\beta}$, errors are introduced:
\begin{align}
    \tilde{Q}(s, \cdot) = \tilde{Q}^{-} (s, \cdot) + \delta(s, \cdot),
\end{align}
\noindent where $\epsilon(s, \cdot)$ denotes the combined approximation error of the Q function and empirical behavior policy. 

Using the approximate implicit Q function in Equation~\ref{eq: q diff perfect}:
\begin{align}
    \eta(\tilde{\pi}) - \eta (\pi_{\beta}) &\approx \int_{s \in \mathcal{S}} \rho_{\pi_{\beta}} (s) (\tilde{Q} (s, \tilde{\pi} (s)) - \tilde{Q} (s, \pi_{\beta} (s)))\, ds
    \\
    &= \int_{s \in \mathcal{S}} \rho_{\pi_{\beta}} (s) (\tilde{Q}^{-}(s, \tilde{\pi} (s)) - \tilde{Q}^{-} (s, \pi_{\beta} (s)) + \delta (s, \tilde{\pi} (s) - \delta (s, \pi_{\beta} (s)))\, ds
    \\
    \label{eq: expanded noisy preferences}
    &= \int_{s \in \mathcal{S}} \rho_{\pi_{\beta}} (s) (\tilde{Q}^{-}(s, \tilde{\pi} (s)) - \tilde{Q}^{-} (s, \pi_{\beta} (s)))\, ds 
    + 
    \int_{s \in \mathcal{S}} \rho_{\pi_{\beta}} (s) (\delta (s, \tilde{\pi} (s) - \delta (s, \pi_{\beta} (s)))\, ds.
\end{align}

When the TVD is bounded, $D_{\text{TV}} (\tilde{Q}^{-} (s, \tilde{\pi}(s)), Q^* (s, \tilde{\pi}(s))) \leq \tilde{\epsilon}$ and $D_{\text{TV}} (\tilde{Q}^{-} (s, \pi_{\beta}(s)), Q^{*} (s, \pi_{\beta}(s))) \leq \epsilon$, then $\forall s \in \mathcal{S}$ we can bound the approximation error:
\begin{align}
    \lvert \delta(s, \tilde{\pi}(s) \rvert &= \lvert \tilde{Q}^{-}(s, \tilde{\pi} (s)) - Q^{*} (s, \tilde{\pi} (s)) \rvert
    \\
    &\leq D_{\text{TV}} (\tilde{Q}^{-} (s, \tilde{\pi}(s)), Q^* (s, \tilde{\pi}(s))) \leq \tilde{\epsilon},
\end{align}
\noindent and 
\begin{align}
    \lvert \delta(s, \pi_{\beta} (s)) \rvert &= \lvert \tilde{Q}^{-} (s, \pi_{\beta}(s)) - Q^{*} (s, \pi_{\beta}(s)) \rvert 
    \\
    &\leq D_{\text{TV}} (\tilde{Q}^{-} (s, \pi_{\beta}(s)), Q^{*} (s, \pi_{\beta}(s))) \leq \epsilon.
\end{align}

Therefore the difference in errors is:
\begin{align}
    \lvert \delta(s, \tilde{\pi} (s)) - \delta (s, \pi_{\beta} (s) \rvert &\leq \lvert \delta(s, \tilde{\pi} (s)) \rvert + \lvert \delta (s, \pi_{\beta} (s)) \rvert
    \\
    &= \tilde{\epsilon} + \epsilon
\end{align}

Let $\rho_{\text{max}} = \sup \{ \rho_{\pi_{\beta}} (s),\ \forall s \in \mathcal{S} \}$. We can find an upper bound for the value of $\rho_{\text{max}}$. When $\pi_{\beta}$ follows that same trajectory every time such that $P(s_t = s') = 1$ then $\rho_{\text{max}} = \sum_{t=0}^{\infty} \gamma^t = \frac{1}{1 - \gamma}$. Next, consider the case when the behavior policy is ``uniform'' in the offline dataset in the sense that it visits all all states with equal probability; then $P(s_t = s) = \frac{1}{N}$ for an offline dataset with $N$ states. The discounted occupancy distribution then becomes $\rho_{\pi_{\beta}} = \sum_{t=0}^{\infty} \gamma^t \frac{1}{N} = \frac{1}{N} \frac{1}{1 - \gamma}$. This is the lower bound as $\rho_{\text{max}}$ as $\int_{s \in \mathcal{S}} \frac{1}{N (1 - \gamma)}\, ds = \frac{1}{1 - \gamma} \geq \rho_{\pi_{\beta}}$, i.e.\ satisfies $\rho_{\text{max}} \geq \rho_{\pi_{\beta}}$. Therefore, we have $\frac{1}{N (1 - \gamma)} \leq \rho_{\text{max}} \leq \frac{1}{1 - \gamma}$. 

Consider the second term in the RHS of Equation~\ref{eq: expanded noisy preferences}:
\begin{align}
    \Delta &= \int_{s \in \mathcal{S}} \rho_{\pi_{\beta}} (s) (\delta (s, \tilde{\pi} (s) - \delta (s, \pi_{\beta} (s)))\, ds
    \\
    &= \rho_{\text{max}}  \int_{s \in \mathcal{S}} \frac{\rho_{\pi_{\beta}} (s)}{\rho_{\text{max}}} (\delta (s, \tilde{\pi} (s) - \delta (s, \pi_{\beta} (s)))\, ds
    \\
    &\leq \rho_{\text{max}}  \int_{s \in \mathcal{S}} \lvert \frac{\rho_{\pi_{\beta}} (s)}{\rho_{\text{max}}} (\delta (s, \tilde{\pi} (s) - \delta (s, \pi_{\beta} (s)))\rvert \, ds.
\end{align}

By definition, we know $\rho_{\pi} (s) \in [0, \frac{1}{1 - \gamma}]$, so $\frac{\rho_{\pi_{\beta}} (s)}{\rho_{\text{max}}} \in [0, 1]$. We use this in the following:
\begin{align}
    \Delta &\leq \rho_{\text{max}}  \int_{s \in \mathcal{S}} \lvert \delta (s, \tilde{\pi} (s) - \delta (s, \pi_{\beta} (s))\rvert \, ds,
\end{align}

\noindent then, using Lemma~\ref{lem: tv distance}:
\begin{align}
    &= 2 \rho_{\text{max}} \max_{s \in \mathcal{S}} \lvert \delta (s, \tilde{\pi} (s) - \delta (s, \pi_{\beta} (s))\rvert
    \\
    &\leq 2 \rho_{\text{max}} (\tilde{\epsilon} + \epsilon).
\end{align}

Finally, plugging this into Equation~\ref{eq: expanded noisy preferences}, we get:
\begin{align}
    \eta(\tilde{\pi}) - \eta (\pi_{\beta}) \leq \int_{s \in \mathcal{S}} \rho_{\pi_{\beta}} (s) (\tilde{Q}^{-}(s, \tilde{\pi} (s)) - \tilde{Q}^{-} (s, \pi_{\beta} (s)))\, ds + 2 \rho_{\text{max}} (\tilde{\epsilon} + \epsilon),
\end{align}
\noindent which yields the empirical:
\begin{align}
    \eta(\tilde{\pi}) - \eta (\pi_{\beta}) \leq \mathbb{E}_{s \sim \mathcal{D}} \left[ \tilde{Q}^{-} (s, \tilde{\pi} (s)) - \tilde{Q}^{-} (s, \pi_{\beta} (s)) \right] + 2 \rho_{\text{max}} (\tilde{\epsilon} + \epsilon).
\end{align}
This completes the proof.

\section{Implementation Details}

\paragraph{EBM} We instantiate the EBM as a neural network following the D2RL architecture \citep{RN924} with hidden layer width 512, with layer normalization \citep{RN606} and spectral normalized fully connected layers \citep{RN888} following \citet{RN885}. We use the training procedure from \citet{RN870} to train the EBM for 200\ 000 gradients steps prior to actor--critic training. 

\paragraph{Critic} We use two critics with two hidden layers of width 256, with layer normalization and clipped double-Q learning (CDQ) \citep{RN714}. 

\paragraph{Policy} We instantiate a Tanh Gaussian policy using an actor network with two hidden layers of width 256. The output layer predicts mean and per-state standard deviation of an isotropic Gaussian distribution.

\paragraph{General} All networks use the GELU nonlinearity \citep{RN463}. We use target Q networks with soft updates at a frequency of 1. We use a batch size of 512 for proprioceptive states and 256 for image-based states. We use $\gamma = 0.99$ for all experiments. For Antmaze tasks, we scale rewards by 100. We use a learning rate of $3e-4$ for all networks with the AdamW optimizer \citep{RN1011}. We evaluate using seeds 1 - 5. 

\paragraph{Entropy} We use soft Bellman updates \citep{RN802} controlled used a hyperparameter $\alpha$. We set $\alpha = 0.2$ in all experiments.

\paragraph{Training + Evaluation} We train the policy for one million gradient steps with periodic evaluation over 10 episodes for Locomotion and 100 for Antmaze. We train an evaluate BPR on an system running Ubuntu 24.04 and we implement our code in Pytorch \citep{RN1014} with acceleration using an Nvidia 1080 Ti. 

\paragraph{V-D4RL} We use a five-layer encoder to map image-states into an embedding. We use two CNNs forming separate encoders, one for the EBM (trained end-to-end with the EBM) and another for the actor--critic networks (again, trained jointly). We evaluate V-D4RL over 10 episodes. 

\subsection{Onestep Experiments}

\paragraph{Onestep BPR} We use two critics with CDQ. For Antmaze tasks we use the discount factor $\gamma = 0.999$. The higher discount factor allows the sparse reward signal to propagate along long trajectories \citep{RN842}. 

\paragraph{Ensemble BPR} Following \citet{RN771}, we use a four-critic ensemble with identical architecture as off-policy BPR. The critics are training independently using $\gamma = 0.999$ and pessimism $\omega = 2.0$. 

\section{Reference Sampling vs.\ Self-Play}

With the reference sampling scheme, we use $\mu_1 = \pi_{\beta} = \mathcal{D}$ and $\mu_2 = \pi$. This directly compares behavior policy actions with samples from the current policy. 

In self-play, both samples are drawn from the current policy using $\mu_1 = \mu_2 = \pi$. 

\begin{figure}[h]
    \centering
    \subfloat[\centering Reward function and behavior policy]{{\includegraphics[width=0.3\textwidth]{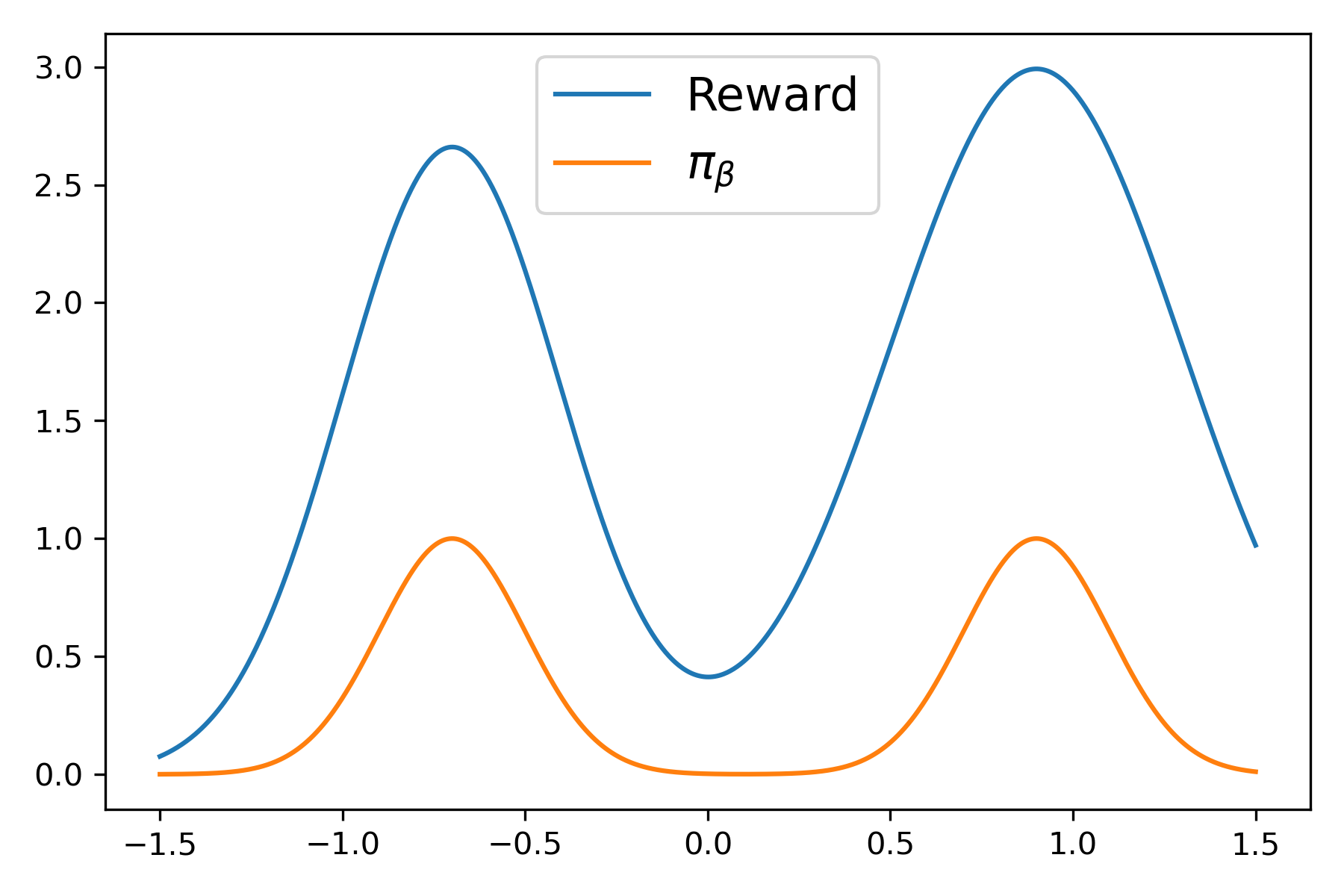} }}%
    \subfloat[\centering Reference sampling fitted policy]{{\includegraphics[width=0.3\textwidth]{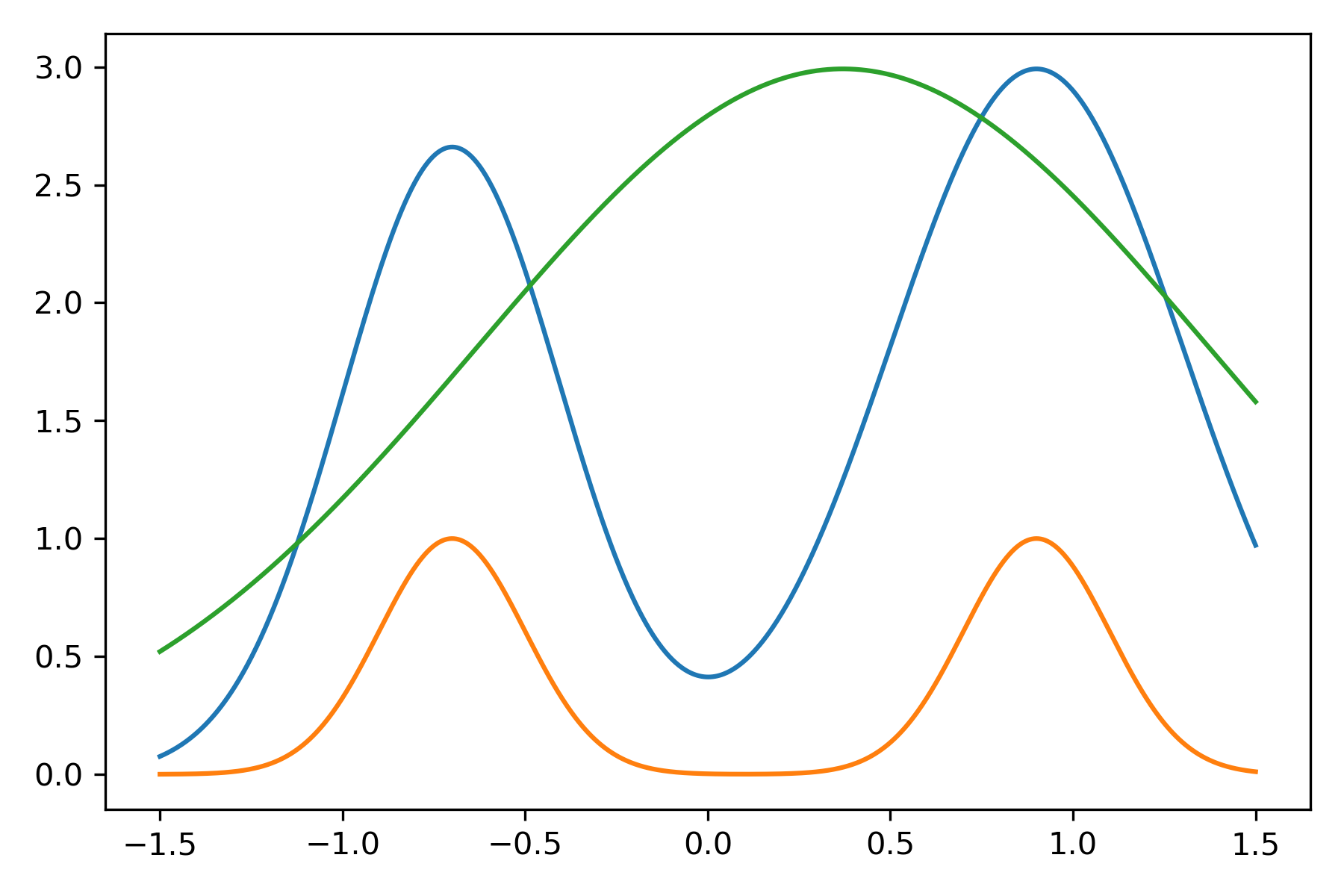} }}%
    \subfloat[\centering Self-play fitted policy]{{\includegraphics[width=0.3\textwidth]{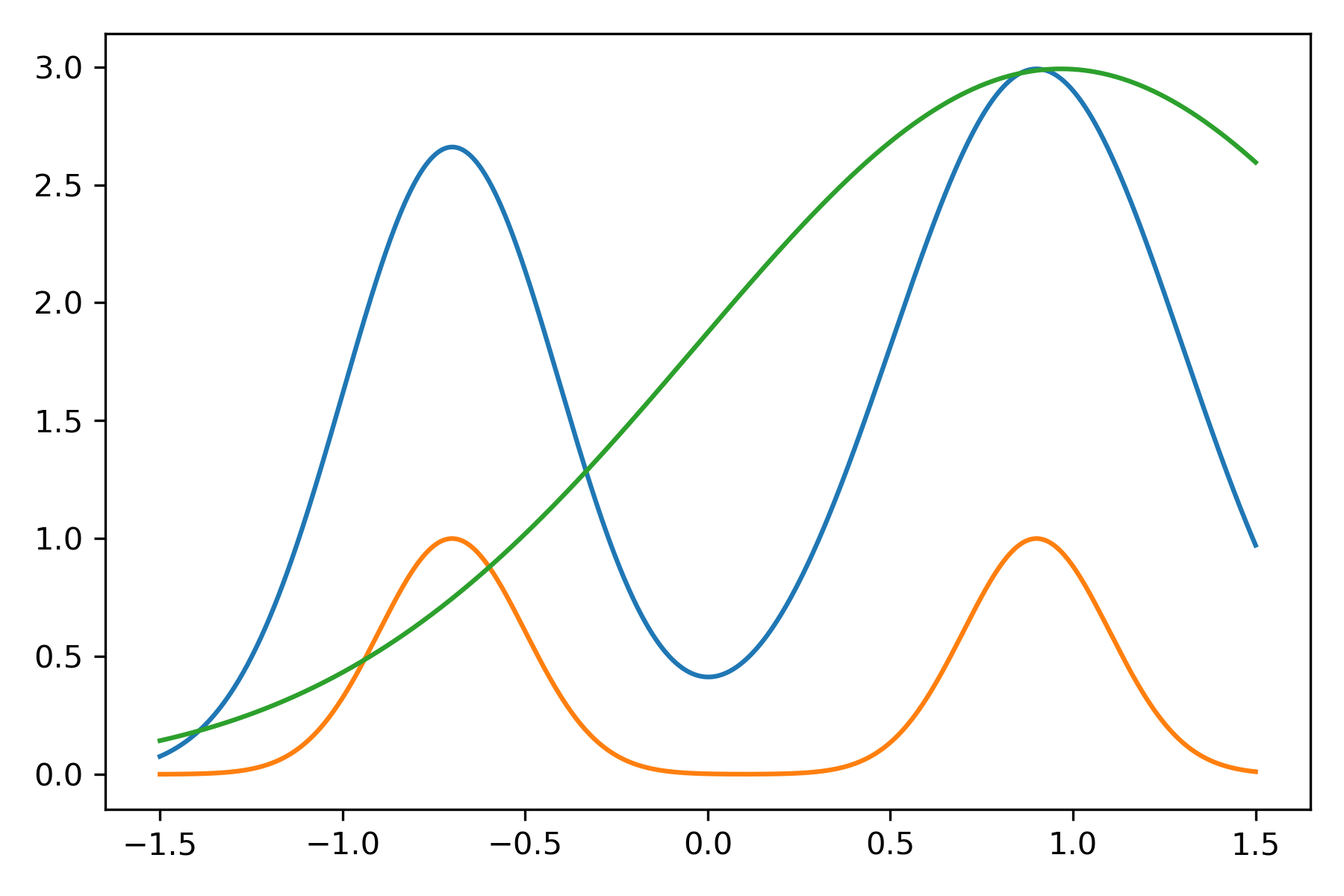} }}%
    \caption{a: Reward function and reference distribution. b: \color{ForestGreen}{Policy}\color{black}\ trained using reference sampling. c: \color{ForestGreen}{Policy}\color{black}\ trained using self-play. Note that all plots have been scaled for clarity and the y-axis corresponds only to the magnitude of the reward function and the x-axis is the action space.}
    \label{fig: reference vs self play}
\end{figure}

We design a toy bandit experiment with a known, reward function and multimodal behavior distribution with one-dimensional actions in [-1, 1]. The offline dataset consists of two actions centered at the modes of the behavior policy and sampled with equal probability for reference sampling. We fit a Gaussian policy using the BPR policy objective for 100\ 000 steps using $\lambda = 1.0$. The plots in Figure~\ref{fig: reference vs self play} show the results of our experiment. In (a), we plot the reward function and behavior policy and fitting a policy using reference sampling in (b) shows that the mean of the policy neither the optimal action or in-support. In contrast, using self-play in (c) produces the optimal policy that is also in-support. 

Reference sampling with $\mu_1 = \mathcal{D}$ introduces an explicit support requirement on the policy, even when the sampled dataset action is suboptimal. The support requirement can centre the policy in OOD regions. This is a known limitation of weighted-BC methods and methods that use forward KL constraints. Prior algorithms address this limitation with extensive hyperparameter tuning that requires online evaluation \citep{RN842,RN706,RN754}.

Self-play allows the policy to fit a mode without suffering support constraints for suboptimal actions, which results in the policy mean both maximizing reward and behavior policy support. This property of self-play is likely a key reason for why hyperparameters generalize well across all datasets in BPR.

While we did experiment with reference sampling in the RL setting, limited evaluation showed that self-play offered more stable, high performance.

\section{Detailed Ensemble BPR Results}

Detailed scores for Ensemble BPR are in Table~\ref{tab: onestep ensemble results} along with Onestep RL (O-RL) \citep{RN882}, CFPI \citep{RN876}, CQL \citep{RN698} (as an off-policy baseline) and Onestep BPR for comparison.

Ensemble BPR slightly underperforms Onestep BPR in the Locomotion tasks but substantially outperforms all other algorithms, including off-policy CQL in the more suboptimal Antmaze datasets. 

$Q_{\text{LCB}}$ ensembles may directly benefit Onestep methods by being less pessimistic at in-sample actions and more pessimisitc (less subject to less exploitation) at OOD actions. 

\begin{table}
    \centering
    \begin{tabular}{lrrrrr}
        \toprule
        Dataset & O-RL & CFPI & CQL & \textbf{Onestep BPR} & \textbf{Ensemble BPR}
        \\
        \midrule
        \texttt{hc-m} & 55.6 & 51.1 & 44.0 & 52.0 $\pm$ 0.8 & 50.7 $\pm$ 0.4
        \\
        \texttt{hp-m} & 83.3 & 86.8 & 58.5 & 96.4 $\pm$ 0.4 & 92.9 $\pm$ 0.2
        \\
        \texttt{w-m} & 85.6 & 88.3 & 72.5 & 89.7 $\pm$ 1.3 & 91.7 $\pm$ 1.9
        \\
        \texttt{hc-m-r} & 41.4 & 44.5 & 45.5 & 51.0 $\pm$ 0.4 & 53.1 $\pm$ 0.2
        \\
        \texttt{h-m-r} & 71.0 & 93.6 & 95.0 & 99.1 $\pm$ 2.3 & 97.4 $\pm$ 1.8
        \\
        \texttt{w-m-r} & 71.6 & 78.2 & 77.2 & 92.0 $\pm$ 0.8 & 87.2 $\pm$ 3.7
        \\
        \texttt{amaze-m-p} & 0.3 & - & 61.2 & 52.7 $\pm$ 10.3 & 67.1 $\pm$ 4.4
        \\
        \texttt{amaze-m-d} & 0.0 & - & 53.7 & 40.0 $\pm$ 7.8 & 71.0 $\pm$ 9.1
        \\
        \texttt{amaze-l-p} & 0.0 & - & 15.8 & 10.4 $\pm$ 2.9 & 33.4 $\pm$ 3.5
        \\
        \texttt{amaze-l-d} & 0.0 & - & 14.9 & 12.7 $\pm$ 1.6 & 27.9 $\pm$ 2.8
        \bottomrule
    \end{tabular}
    \caption{Scores for Ensemble BPR with Onestep RL, Onestep CFPI, CQL and Onestep BPR. We evaluate on non-expert Locomotion and \texttt{medium} and \texttt{large} Antmaze (\texttt{amaze}) datasets. The authors of CFPI do not report Onestep results for Antmaze. We report 1 SD for Onestep and Ensemble BPR.}
    \label{tab: onestep ensemble results}
\end{table}

\section{Ablation Experiments}

\begin{figure}[h]
    \centering
    \includegraphics[width=0.9\textwidth]{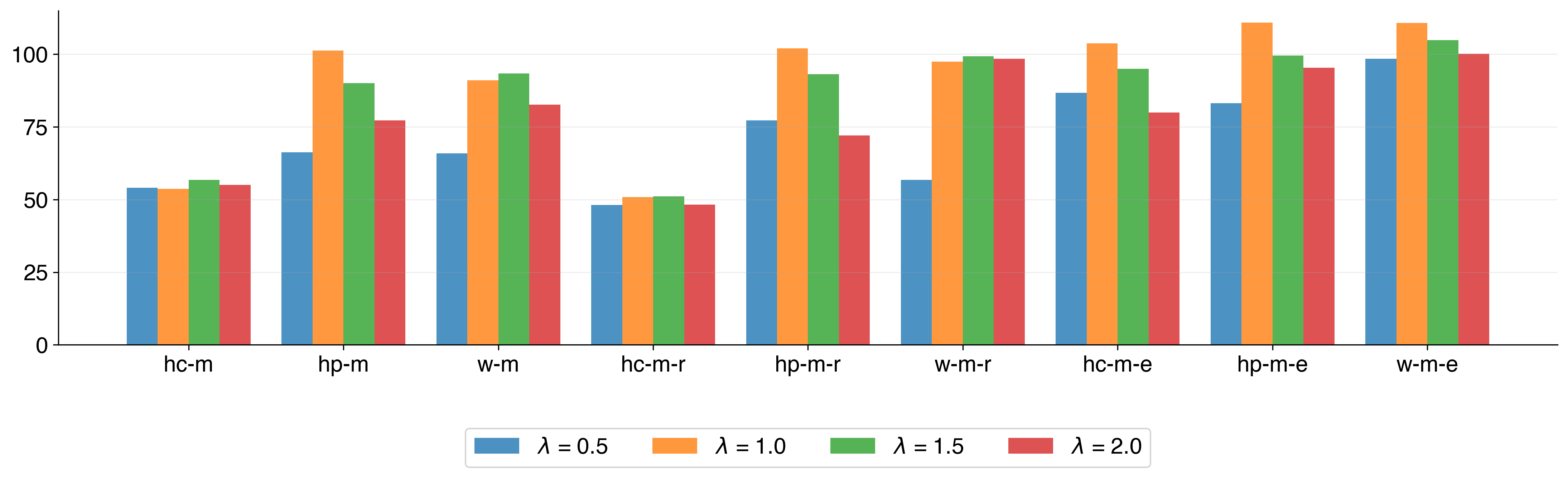} 
    \caption{Ablations for different values of $\lambda$. We use $\lambda = 1.0$ for results reported in the main paper.}
    \label{fig: lambda ablations}
\end{figure}

We examine sensitivity to the hyperparameter $\lambda$ in Figure~\ref{fig: lambda ablations}. We see that $\lambda = 1.0$ performs consistently well, usually outperforming all other values. 

\newpage
\bibliography{mybib}